\documentclass[conference]{IEEEtran}
\IEEEoverridecommandlockouts
\usepackage{cite}
\usepackage{amsmath,amssymb,amsfonts}
\usepackage{algorithmic}
\usepackage{graphicx}
\usepackage{textcomp}
\usepackage{xcolor}
\usepackage[colorlinks, urlcolor=black, linkcolor=blue, citecolor=blue]{hyperref}
\def\BibTeX{{\rm B\kern-.05em{\sc i\kern-.025em b}\kern-.08em
    T\kern-.1667em\lower.7ex\hbox{E}\kern-.125emX}}
\begin{document}

\title{Towards Cross-Scale Attention and Surface Supervision for Fractured Bone Segmentation in CT}

\author{\IEEEauthorblockN{Yu Zhou, Xiahao Zou, Yi Wang$^{*}$
\thanks{$*$ Corresponding author}
\thanks{This work was supported in part by the Guangdong-Hong Kong Joint Funding for Technology and Innovation under Grant 2023A0505010021,
in part by the National Natural Science Foundation of China under Grant 62071305,
and in part by the Guangdong Basic and Applied Basic Research Foundation under Grant 2022A1515011241.}	
}
\IEEEauthorblockA{
Smart Medical Imaging, Learning and Engineering (SMILE) Lab,\\
Medical UltraSound Image Computing (MUSIC) Lab,
\\School of Biomedical Engineering, Shenzhen University Medical School, Shenzhen University, Shenzhen, 518060, China}
}

\maketitle

\begin{abstract}
Bone segmentation is an essential step for the preoperative planning of fracture trauma surgery.
The automated segmentation of fractured bone from computed tomography (CT) scans remains challenging, due to the large differences of fractures in position and morphology, and also the inherent anatomical characteristics of different bone structures.
To alleviate these issues, we propose a cross-scale attention mechanism as well as a surface supervision strategy for fractured bone segmentation in CT.
Specifically, a cross-scale attention mechanism is introduced to effectively aggregate the features among different scales to provide more powerful fracture representation.
Moreover, a surface supervision strategy is employed, which explicitly constrains the network to pay more attention to the bone boundary.
The efficacy of the proposed method is evaluated on a public dataset containing CT scans with hip fractures.
The evaluation metrics are Dice similarity coefficient (DSC), average symmetric surface distance (ASSD), and Hausdorff distance (95HD).
The proposed method achieves an average DSC of 93.36\%, ASSD of 0.85mm, 95HD of 7.51mm.
Our method offers an effective fracture segmentation approach for the pelvic CT examinations, and has the potential to be used for improving the segmentation performance of other types of fractures.
Our code is publicly available at \textit{\url{https://github.com/ZhouyuPOP/FracSeg-Net}}.
\end{abstract}


\section{Introduction}
China has one of the fastest aging populations in the world, and the number of fracture patients is increasing year by year~\cite{REN2019143}.
As a result of this demographic shift, certain types of fractures are becoming more prevalent.
In middle-aged and elderly individuals, hip fractures can be caused by mild violence because of osteoporosis-related bone changes, which accounts for a large proportion of age-related fractures.
As the aging trend accelerates, the prevalence of fractures is inevitable to increase further.

Adequate evaluation and diagnosis of fractures is critical to selecting the optimal treatment strategy to minimize disability and other complications in patients.
However, the clinical diagnosis and treatment of fractures are difficult due to the different types of fractures, the patient's own circumstances and the hospital qualification.
In order to obtain good operative effect of fractures, it is necessary to make detailed preoperative planning according to the specific situation of fractures.
As for the patient-specific preoperative planning, 
automated and accurate bone segmentation from computed tomography (CT) scans is an essential step to facilitate the quantitative interpretation of fractures.


\begin{figure}[t]
	\begin{center}
		\includegraphics[width=\linewidth]{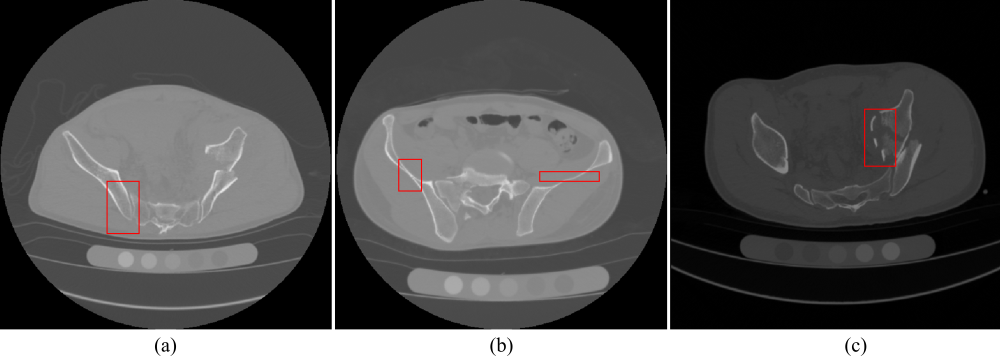}
	\end{center}
	\caption{Illustration of bone CT slices from three subjects. It can be observed that due to the inherent anatomical characteristics of the bone structures (the adjacent region of multi-bones in (a) and the narrow and thin regions in (b)), and the various types of fractures in (c), the automated segmentation remains challenging.}
	\label{fig:challenge}
\end{figure}

Although lots of medical image segmentation methods have been proposed and further employed in various anatomical structures including brain~\cite{zeng2023reciprocal}, breast~\cite{zhong2023simple}, abdomen~\cite{YANG2024122024}, prostate~\cite{wangMICCAI} and so on, accurate fracture segmentation in CT scans is still very challenging, as shown in Fig.~\ref{fig:challenge}.
First of all, the inherent anatomical characteristics of the bone pose difficulties in accurate segmentation.
The boundaries of bone structures might not easy to be delineated in the adjacent region of multi-bones (see Fig.~\ref{fig:challenge}(a)), and the narrow and thin regions (see Fig.~\ref{fig:challenge}(b)).
Secondly, fractures have large differences in position and morphology.
Moreover, fractures frequently expose cancellous bone, which shares similar intensity to surrounding soft tissue (see Fig.~\ref{fig:challenge}(c)), and therefore resulting in challenges in fracture segmentation.


In this study, we propose a cross-scale attention mechanism as well as a surface supervision strategy to boost the performance of conventional segmentation network for fractured bone CT scans.
First of all, a cross-scale attention mechanism is introduced to effectively capture and fuse the features among different scales to provide more powerful fracture representation.
Additionally, considering the primary feature of the bone in CT scans is the boundary information, we devise a surface loss to supervise the training of the segmentation network.
The proposed surface supervision explicitly constrains the network to pay attention to the bone boundary and therefore providing more accurate bone segmentation.
We evaluate the proposed method on a publicly available dataset, which contains 103 pelvic CT scans with hip fractures.
Experimental results demonstrate that our method achieves satisfactory fracture segmentation performance.


\section{Related Work}
\label{sec:literature}
There are several studies aimed at bone segmentation using traditional image processing methods.
Shadid~\textit{et al}.~\cite{work9} used a probability-based variation of the watershed transform.
Each voxel in the 3D images was classified as marker, bone, or background, then a set of probability distributions was generated to segment the fractured bone.
Kr{\v{c}}ah~\textit{et al}.~\cite{work10} proposed a fully-automatic method for segmenting the femur in CT, which employed the method of graph-cuts and a bone boundary enhancement filter to analyze the second-order local structure instead of the conventional statistical shape and intensity modeling method.
Inspired by~\cite{work10}, Besler~\textit{et al}.~\cite{work11} modified its scheme to a semi-automatic version that allowed the observer to sparsely annotate the images to provide guidance for the graph-cuts segmentation algorithm.

In recent years, a large number of related studies utilized deep convolutional neural networks (CNNs) as the solution for medical image processing.
Mu~\textit{et al}.~\cite{work2} designed a multi-task U-net to detect and segment fractured calcaneus in radiographs.
By applying the multi-task supervision, the detection and segmentation performance can promote each other.
Chen~\textit{et al}.~\cite{work3} attempted to segment the wrist bone.
They first extracted bone region-of-interest (ROI) from radiographs and then segmented the wrist bone through a variant of U-net by adding the attention module and residual module in the network.
Rehman~\textit{et al}.~\cite{work4} combined traditional region-based level set method with deep learning framework to predict the shape of vertebral bones, which showed satisfactory performance on cases without fractures but not accurate enough for fractured cases.
Yang~\textit{et al}.~\cite{work17} proposed a cascaded architecture combining 2D U-net and Mask-RCNN to recognize and segment individual fragments of complex intertrochanteric fractures in CT images.
However, such 2D CNN did not utilize the 3D spatial information sufficiently.
Chen~\textit{et al}.~\cite{work6} mentioned that the weak bone boundaries and the narrowness of joint space are main obstacles in femur segmentation task, so it is of great importance to enhance the boundary.
They designed a multi-task 3D network to detect the bone boundary and segment the femur.
Chen~\textit{et al}.~\cite{chen2021csr} introduced an attention mechanism and the residual connection into the encoder-decoder structure to segment scaphoid fractures.

\section{Materials and Methods}
\label{sec:method}
\subsection{Materials}
To validate the efficacy of the proposed method, experiments are conducted on a publicly available dataset containing pelvic CT scans with hip fractures.
This dataset CTPelvic1K~\cite{work8} consists of seven sub-datasets.
Most sub-datasets are collected mainly focusing on organ and tumor segmentation tasks.
We choose the sub-dataset6 in CTPelvic1K as our experimental dataset, which contains the preoperative CT scans of patients suffered from hip bone fractures.
All the ground-truth segmentation masks of the hip bone fractures are provided.
The experimental dataset includes 103 pelvic CT scans.
The original volume size ranges from 512$\times$512$\times$294 to 512$\times$512$\times$388, and the spacing ranges from 0.6399$\times$0.6399$\times$0.7990 mm$^3$ to 1.129$\times$1.129$\times$0.8010 mm$^3$.
%
For the original CT volumes, a coarse filter is applied to filter out the voxels with CT value below -200 HU or over 800 HU.
Then, a windowing and centering operation is employed to enhance and normalize the CT volumes.
All the volumes are resampled to the spacing of 0.8$\times$0.8$\times$0.8 mm$^3$.
The public dataset is randomly split into training set and testing set with the ratio of 8:2.
Finally, 83 volumes are used to construct the segmentation network and 20 volumes are employed to validate the performance.

\subsection{The Proposed Fracture Segmentation Network}
\subsubsection{Network Architecture}
\begin{figure}[t]
	\begin{center}
		\includegraphics[width=\linewidth]{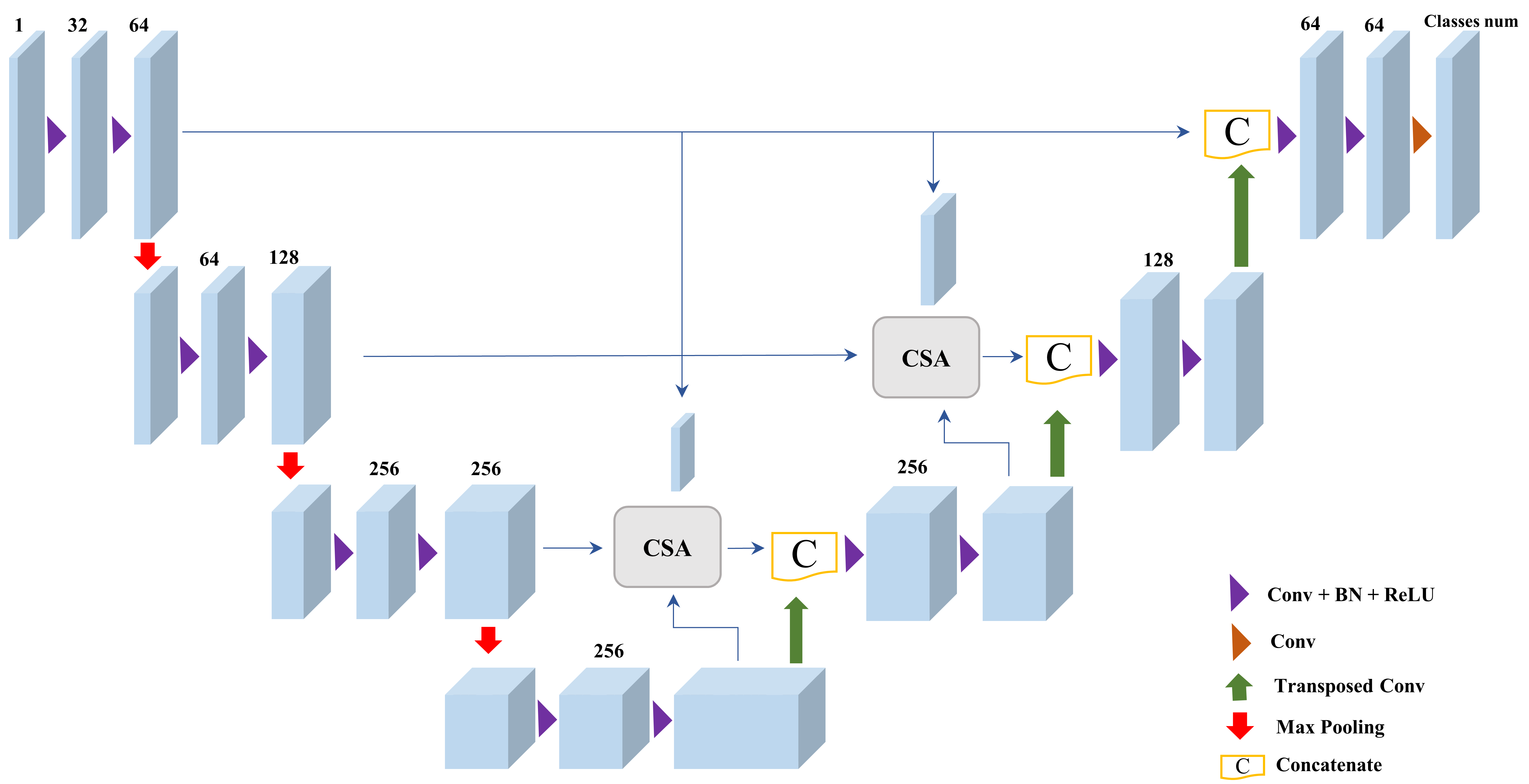}
	\end{center}
	\caption{The network architecture of the fracture segmentation model. The backbone is based on 3D U-net~\cite{work14}. The cross-scale attention (CSA) modules are used to enhance the feature representation. Conv: convolution; BN: batch normalization; ReLU: rectified linear unit.}
	\label{fig:network}
\end{figure}

The backbone architecture is based on the 3D U-net~\cite{work14}, which addresses the issue of three-dimension medical image segmentation.
The detailed architecture of our segmentation network is illustrated in Fig.~\ref{fig:network}, which consists of four encoder blocks and three decoder blocks in total.
The encoder path is used to analyze and extract features from the given input CT patches, where the decoder path creates semantic segmented maps by skip connection operations.
Each encoder block consists of two consecutive 3$\times$3$\times$3 convolutions and a 2$\times$2$\times$2 max pooling operation with stride of 2 in each dimension.
Each convolution is followed by a batch normalization (BN) and a rectified linear unit (ReLU).
On the other hand, each decoder block consists of two consecutive 3$\times$3$\times$3 convolutions and a 3$\times$3$\times$3 transposed convolution with stride of 2 and padding of 1 in each dimension, also, BN and ReLU are followed sequentially after convolution.
Finally, the soft-max activation is applied to obtain segmented predictions.

It is worth noting that unlike the conventional skip connection in~\cite{work14} which directly fuses the encoding feature maps with the same resolution feature maps generated by the decoders, we design the cross-scale attention (CSA) module to generate attention maps through different scales of feature maps.
In particular, we apply the CSA modules at the second and the third layers, thus strengthening the ability of capturing more powerful features and effectively fusing the highly semantic information as well as detailed information.
The detailed structure of the CSA module is described in the following section.

\begin{figure}[t]
	\begin{center}
		\includegraphics[width=\linewidth]{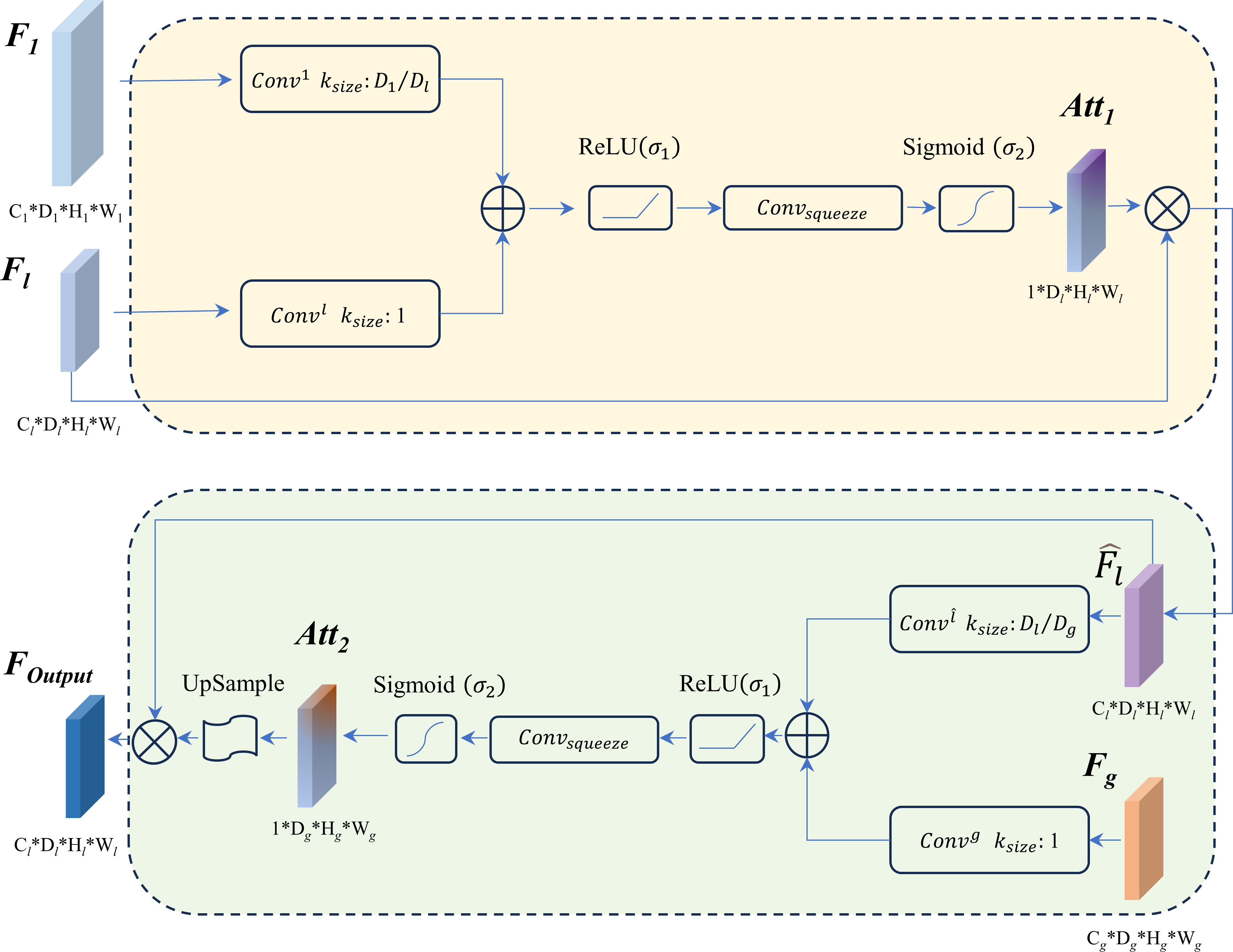}
	\end{center}
	\caption{
		Detailed structure of the proposed cross-scale attention (CSA) module.
		The W, H, D, C denote the width, height, depth and channel number of the feature map, respectively.
		The $Conv$ is the convolution operation with the kernel size $k_{size}$.
	}
	\label{fig:CSA}
\end{figure}

\subsubsection{Cross-Scale Attention}
Various attention mechanisms \cite{8698868, yang2024recurrent, lin2021variance, huang2023joint} have been proposed to boost the performance of deep neural networks.
In this study, we devise two successive attention operations to progressively enhance feature responses in specific object regions while simultaneously preserving detailed information.
As shown in Fig.~\ref{fig:CSA}, the CSA module takes the $1^{st}$ layer feature maps of the encoder path $F_{1}$, the $l^{th}$ layer feature maps of the encoder path $F_{l}$, and the $g^{th}$ layer feature maps of the decoder path $F_{g}$ as inputs.
The $F_{1}$ is utilized for each voxel in feature map $F_{l}$ to determine detailed regions.
This is crucial for tasks such as fractured bone segmentation, where detailed information tends to weaken in subsequent layers.
The attention coefficients derived from detailed information, denoted as $Att_{1}$, can be formulated as:
\begin{equation}
Att_{1} = \sigma_{2}(Conv_{squeeze}(\sigma_{1} (Conv^{l}(F_{l})+Conv^{1}(F_{1})))),
\label{Att1}
\end{equation}
where $Conv_{squeeze}$ is a convolution operation to squeeze the information in the channel dimension~\cite{Hu2018CVPR}.
By element-wise multiplication of $F_{l}$ and $Att_{1}$, $\hat{F_{l}}$ is obtained, enriching the low-level features considerably.
The $\hat{F_{l}}$ then queries representative features from the sparse but semantically meaningful feature maps $F_{g}$ in a similar manner.
The calculation of $Att_{2}$ is as follows:
\begin{equation}
Att_{2} = \sigma_{2}(Conv_{squeeze}(\sigma_{1} (Conv^{g}(F_{g})+Conv^{\hat{l} }(\hat{F_{l}})))).
\label{Att2}
\end{equation}
Finally, by upsampling $Att_{2}$ and performing an element-wise multiplication with $\hat{F_{l}}$, the output of the CSA module $F_{output}$ is generated.
So far, $F_{output}$ aggregates information across multiple scales.

\subsubsection{Surface Supervision}
Conventional segmentation networks often use region-based Dice loss~\cite{diceloss, wang20203d, wang2019deeply} to supervise the network training.
However, the Dice loss mainly evaluates the global regional similarity and therefore may neglect the learning of the local boundary/surface.
To mitigate this issue, we explicitly supervise the 3D surface similarity between the ground-truth bone surface and the network's regional soft-max probability outputs~\cite{kervadec2019boundary}.
By doing so, the segmentation network pays more attention to the bone boundary thus providing more accurate bone segmentation.
The surface loss $\mathcal{L}_{surface}$ can be defined as follows:
\begin{equation}
\mathcal{L}_{surface} = \sum_{q\in\Omega}^{}  \phi_G(q) s(q),
\label{surface_loss}
\end{equation}
where $q$ is the voxel in the volume $\Omega$,
$s(q)$ is the soft-max probability output of $q$,
and $\phi_G(q)$ denotes the distance from $q$ to the ground-truth bone surface $\partial G$, which can be calculated as:
\begin{equation}
\phi_G(q)=\begin{cases}-D_G\left ( q \right ) ,   q\in G
\\D_G\left ( q \right ) ,   otherwise
\end{cases}
\end{equation}
where
\begin{equation}
D_G(q) = \min_{p\in \partial G} \left \|q-p  \right \|,
\end{equation}
where $G$ is the ground-truth bone segmentation mask.
The absolute value of $\phi_G(q)$ is the Euclidean distance from $q$ to $\partial G$, with bigger magnitude for the longer distance from the bone surface.
The sign of $\phi_G(q)$ is negative if voxel $q$ is within the bone region, and positive when $q$ is outside the bone region.

In order to train the network more effectively, the conventional region-based Dice loss $\mathcal{L}_{dice}$~\cite{diceloss} is also used.
The Dice loss aims to maximize the overlap regions between the ground-truth and predicted segmentation.
Therefore, the total training loss is:
\begin{equation}
\mathcal{L}_{train} = \mathcal{L}_{surface} + \lambda\mathcal{L}_{dice},
\label{Loss_total}
\end{equation}
where $\lambda$ is a weighting coefficient.

\subsubsection{Implementations}
The experimental environment is as follows: NVIDIA Tesla V100 GPU, Linux operating system, Py-torch deep learning framework + python3.7.
To train the network, Adam optimizer is selected and the initial learning rate is 0.01, beta1 and beta2 is 0.5 and 0.999, respectively.
The batch size is set to 6 due to the GPU memory limitation.
The parameter $\lambda$ is set to 1.
During the training procedure, the learning rate changes with the increase of training epochs following the Cosine Annealing learning scheduler.
The maximum training epoch is set to 150.
The code is publicly available at \textit{\url{https://github.com/ZhouyuPOP/FracSeg-Net}}.

\subsection{Comparison and Evaluation}
We compare our method with several established segmentation approaches, including the graph-cut method~\cite{work10}, the conventional 3D U-net model, and the Transformer-based models (UNETR~\cite{unetr} and Swin-UNETR~\cite{swin}).
The graph-cut method~\cite{work10} is a fully automated method that segments bone structures in CT images by using several image processing techniques.
We generate the results from the public graph-cut codes provided by the authors using the recommended parameter setting.
Both UNETR~\cite{unetr} and Swin-UNETR~\cite{swin} are built upon the vision Transformer~\cite{vit}, which incorporate the self-attention mechanism into the U-shaped architecture.
The 3D U-net model, along with the two Transformer-based models, are trained using the conventional Dice loss.
We adjust their training parameters to obtain the best segmentation results.
We further separately apply the surface supervision and cross-scale attention for the U-net model, denoted as U-net+$\mathcal{L}_{surface}$ and U-net+CSA, respectively.
These two models are regarded as the ablation models of our method.

The evaluation metrics are Dice similarity coefficient (DSC), average symmetric surface distance (ASSD), and symmetric 95\% Hausdorff distance (95HD).
The DSC is used to assess the relative volumetric overlap between the segmented regions and the ground-truth masks.
The ASSD is employed to evaluate the distance between the segmented surfaces and the ground-truth surfaces.
The 95HD is similar to ASSD but more sensitive to the localized disagreement as it determines the 95th percentile of all calculated Hausdorff distances.
A better segmentation shall have smaller ASSD and 95HD, and larger value of DSC metric.

\begin{table}[t]
	\centering
	\caption{The numerical segmentation results of different methods for hip fractures (Mean$\pm$SD, best results are highlighted in bold). ``$*$'' indicates the results are statistically different with ours (Wilcoxon tests, p $\textless$ 0.05).}
	\label{tab:resultsHF}
	\setlength{\tabcolsep}{2.5mm}{
		\begin{tabular}{l|c|c|c}
			\hline
			Methods & DSC (\%) & ASSD (mm) & 95HD (mm) \\
			\hline
			Graph-cut~\cite{work10} & 88.39$\pm$11.43* & 1.84$\pm$2.44* & 17.38$\pm$23.26* \\
			UNETR~\cite{unetr} & 77.24$\pm$6.97* & 6.89$\pm$2.59* & 69.36$\pm$19.29* \\
			Swin-UNETR~\cite{swin} & 78.48$\pm$3.33* & 10.99$\pm$2.65* & 78.75$\pm$14.48* \\
			U-net & 77.64$\pm$4.63* & 8.97$\pm$2.75* & 72.53$\pm$17.23* \\
			U-net+$\mathcal{L}_{surface}$ & 92.56$\pm$2.54* & 2.32$\pm$2.09* & 28.99$\pm$35.01* \\
			U-net+CSA & \textbf{93.73$\pm$1.69} & 2.19$\pm$2.49* & 22.19$\pm$38.62* \\
			Ours & 93.36$\pm$1.43 & \textbf{0.85$\pm$0.94} & \textbf{7.51$\pm$25.02} \\
			\hline
		\end{tabular}%
	}
\end{table}

\section{Experimental Results}
\label{sec:results}
The numerical segmentation results are listed in Table~\ref{tab:resultsHF}.
It can be observed that our method consistently outperforms graph-cut and deep-learning-based methods with respect to all evaluation metrics.
Specifically, our method attains an average DSC of 93.36\%, ASSD of 0.85mm, 95HD of 7.51mm on the dataset of hip fractures.
To investigate the statistical significance of our method over other comparison methods on each metrics, the Wilcoxon signed-rank test is analyzed.
By observing Table~\ref{tab:resultsHF}, it can be concluded that the null hypotheses for all comparing pairs on all metrics are not accepted at the 0.05 level.
As a result, our method can be regarded as significantly better than other comparison methods.
By employing the cross-scale attention and surface supervision, our network effectively boosts the performance of the conventional U-net model in the application of fracture segmentation.

Table~\ref{tab:resultsHF} also shows the ablation results.
Compared with the two ablation models U-net+$\mathcal{L}_{surface}$ and U-net+CSA, 
our method achieves similar region-based DSC results but with significantly better boundary-based ASSD and 95HD results.
This demonstrates that either cross-scale attention or surface supervision enhances the performance of U-net model, while applying both two, the performance can be further boosted particularly to generate better local boundaries.

\begin{figure}[t]
	\begin{center}
		\includegraphics[width=\linewidth]{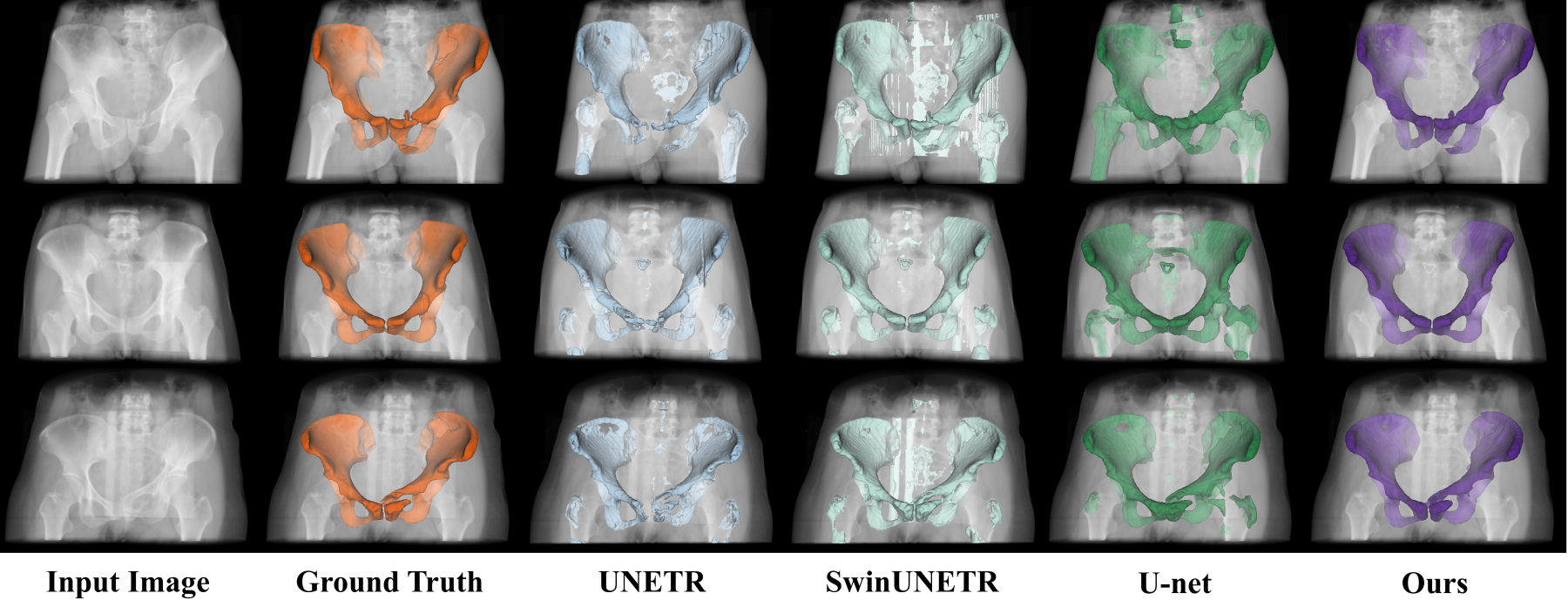}
	\end{center}
	\caption{3D visualization of the segmentation results from different methods.}
	\label{fig:3dresultsHF}
\end{figure}

\begin{figure}[t]
	\begin{center}
		\includegraphics[width=\linewidth]{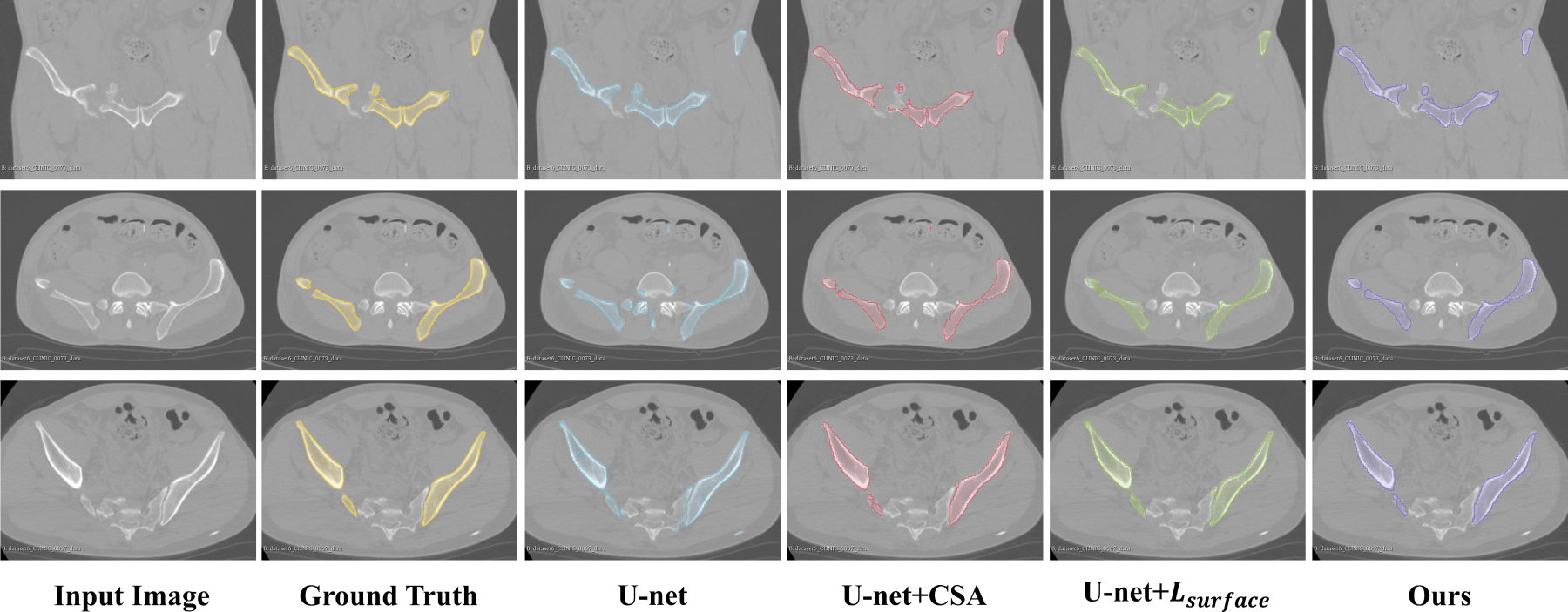}
	\end{center}
	\caption{2D visualization of the segmentation results on hip fractures.}
	\label{fig:2dresultsHF}
\end{figure}

Fig.~\ref{fig:3dresultsHF} shows the 3D visualization of segmentation results from different deep-learning-based methods.
It is obvious that our method provides more accurate segmentation results.
Both UNETR and Swin-UNETR fail to produce smooth and accurate segmented surfaces.
Conventional U-net cannot handle complicated fractures, particularly prone to producing false-positive predictions in irrelevant background regions, while our attention-enhanced and surface-supervised method generates more similar segmented results to the ground-truths.
Fig.~\ref{fig:2dresultsHF} further visualizes some detailed 2D comparisons.
Our method generates most similar segmented boundaries to the ground-truths.

\section{Conclusion}
\label{sec:conc}
This study presents an accurate and automated method for the application of fracture segmentation in CT scans.
The key motivation is to leverage the primary feature of bone structures, i.e., the detailed boundary information, to boost the performance of the segmentation network.
To this end, we design a cross-scale attention mechanism to fuse the shallow features with semantic features and therefore enriching the detailed low-level representation.
In addition, we employ a surface supervision to explicitly evaluate the similarity between the predicted boundaries and ground-truth boundaries.
Experiments on CT scans with hip fractures demonstrate the effectiveness of the devised method.
Compared with conventional image processing method and existing deep-learning-based methods, our proposed attention-enhanced and surface-supervised network outcompetes their performance by large margins.
In addition, the proposed attention mechanism and surface supervision strategy have the potential to be applied for other segmentation backbones and applications.


\bibliographystyle{IEEEtran}
\bibliography{scholar}

\begin{thebibliography}{10}
\providecommand{\url}[1]{#1}
\csname url@samestyle\endcsname
\providecommand{\newblock}{\relax}
\providecommand{\bibinfo}[2]{#2}
\providecommand{\BIBentrySTDinterwordspacing}{\spaceskip=0pt\relax}
\providecommand{\BIBentryALTinterwordstretchfactor}{4}
\providecommand{\BIBentryALTinterwordspacing}{\spaceskip=\fontdimen2\font plus
\BIBentryALTinterwordstretchfactor\fontdimen3\font minus
  \fontdimen4\font\relax}
\providecommand{\BIBforeignlanguage}[2]{{%
\expandafter\ifx\csname l@#1\endcsname\relax
\typeout{** WARNING: IEEEtran.bst: No hyphenation pattern has been}%
\typeout{** loaded for the language `#1'. Using the pattern for}%
\typeout{** the default language instead.}%
\else
\language=\csname l@#1\endcsname
\fi
#2}}
\providecommand{\BIBdecl}{\relax}
\BIBdecl

\bibitem{REN2019143}
Y.~Ren, J.~Hu, B.~Lu, W.~Zhou, and B.~Tan, ``Prevalence and risk factors of hip
  fracture in a middle-aged and older chinese population,'' \emph{Bone}, vol.
  122, pp. 143--149, 2019.

\bibitem{zeng2023reciprocal}
X.~Zeng, R.~Huang, Y.~Zhong, Z.~Xu, Z.~Liu, and Y.~Wang, ``A reciprocal
  learning strategy for semisupervised medical image segmentation,''
  \emph{Medical Physics}, vol.~50, no.~1, pp. 163--177, 2023.

\bibitem{zhong2023simple}
Y.~Zhong and Y.~Wang, ``Sim{PL}e: Similarity-aware propagation learning for
  weakly-supervised breast cancer segmentation in {DCE}-{MRI},'' in
  \emph{Medical Image Computing and Computer-Assisted Intervention--MICCAI
  2023}.\hskip 1em plus 0.5em minus 0.4em\relax Springer, 2023, pp. 567--577.

\bibitem{YANG2024122024}
Z.~Yang, D.~Lin, D.~Ni, and Y.~Wang, ``Non-iterative scribble-supervised
  learning with pacing pseudo-masks for medical image segmentation,''
  \emph{Expert Systems with Applications}, vol. 238, p. 122024, 2024.

\bibitem{wangMICCAI}
Y.~Wang, Z.~Deng, X.~Hu, L.~Zhu, X.~Yang, X.~Xu, P.-A. Heng, and D.~Ni, ``Deep
  attentional features for prostate segmentation in ultrasound,'' in
  \emph{Medical Image Computing and Computer Assisted Intervention -- MICCAI
  2018}, 2018, pp. 523--530.

\bibitem{work9}
W.~G. Shadid and A.~Willis, ``Bone fragment segmentation from 3{D} {CT}
  imagery,'' \emph{Computerized Medical Imaging and Graphics}, vol.~66, pp.
  14--27, 2018.

\bibitem{work10}
M.~Kr{\v{c}}ah, G.~Sz{\'e}kely, and R.~Blanc, ``Fully automatic and fast
  segmentation of the femur bone from 3{D}-{CT} images with no shape prior,''
  in \emph{2011 IEEE international symposium on biomedical imaging: from nano
  to macro}.\hskip 1em plus 0.5em minus 0.4em\relax IEEE, 2011, pp. 2087--2090.

\bibitem{work11}
B.~A. Besler, A.~S. Michalski, M.~T. Kuczynski, A.~Abid, N.~D. Forkert, and
  S.~K. Boyd, ``Bone and joint enhancement filtering: Application to proximal
  femur segmentation from uncalibrated computed tomography datasets,''
  \emph{Medical image analysis}, vol.~67, p. 101887, 2021.

\bibitem{work2}
Y.~Mu, D.~Xue, J.~Guo, H.~Xu, W.~Wang, and H.~Li, ``Automatic calcaneus
  fracture identification and segmentation using a multi-task u-net,'' in
  \emph{2020 5th International Conference on Communication, Image and Signal
  Processing (CCISP)}.\hskip 1em plus 0.5em minus 0.4em\relax IEEE, 2020, pp.
  140--144.

\bibitem{work3}
L.~Chen, X.~Zhou, M.~Wang, J.~Qiu, M.~Cao, and K.~Mao, ``Aru-net: Research and
  application for wrist reference bone segmentation,'' \emph{IEEE Access},
  vol.~7, pp. 166\,930--166\,938, 2019.

\bibitem{work4}
F.~Rehman, S.~I. Ali~Shah, M.~N. Riaz, and S.~O. Gilani, ``A region-based deep
  level set formulation for vertebral bone segmentation of osteoporotic
  fractures,'' \emph{Journal of digital imaging}, vol.~33, pp. 191--203, 2020.

\bibitem{work17}
L.~Yang, S.~Gao, P.~Li, J.~Shi, and F.~Zhou, ``Recognition and segmentation of
  individual bone fragments with a deep learning approach in ct scans of
  complex intertrochanteric fractures: A retrospective study,'' \emph{Journal
  of Digital Imaging}, vol.~35, no.~6, pp. 1681--1689, 2022.

\bibitem{work6}
F.~Chen, J.~Liu, Z.~Zhao, M.~Zhu, and H.~Liao, ``Three-dimensional
  feature-enhanced network for automatic femur segmentation,'' \emph{IEEE
  journal of biomedical and health informatics}, vol.~23, no.~1, pp. 243--252,
  2017.

\bibitem{chen2021csr}
C.~Chen, B.~Liu, K.~Zhou, W.~He, F.~Yan, Z.~Wang, and R.~Xiao, ``C{SR}-{N}et:
  Cross-scale residual network for multi-objective scaphoid fracture
  segmentation,'' \emph{Computers in Biology and Medicine}, vol. 137, p.
  104776, 2021.

\bibitem{work8}
P.~Liu, H.~Han, Y.~Du, H.~Zhu, Y.~Li, F.~Gu, H.~Xiao, J.~Li, C.~Zhao, L.~Xiao,
  X.~Wu, and S.~K. Zhou, ``Deep learning to segment pelvic bones: large-scale
  {CT} datasets and baseline models,'' \emph{International Journal of Computer
  Assisted Radiology and Surgery}, vol.~16, pp. 749--756, 2021.

\bibitem{work14}
{\"O}.~{\c{C}}i{\c{c}}ek, A.~Abdulkadir, S.~S. Lienkamp, T.~Brox, and
  O.~Ronneberger, ``3{D} {U}-net: learning dense volumetric segmentation from
  sparse annotation,'' in \emph{Medical Image Computing and Computer-Assisted
  Intervention--MICCAI 2016}.\hskip 1em plus 0.5em minus 0.4em\relax Springer,
  2016, pp. 424--432.

\bibitem{8698868}
Y.~Wang, H.~Dou, X.~Hu, L.~Zhu, X.~Yang, M.~Xu, J.~Qin, P.-A. Heng, T.~Wang,
  and D.~Ni, ``Deep attentive features for prostate segmentation in 3{D}
  transrectal ultrasound,'' \emph{IEEE Transactions on Medical Imaging},
  vol.~38, no.~12, pp. 2768--2778, 2019.

\bibitem{yang2024recurrent}
Z.~Yang, D.~Lin, D.~Ni, and Y.~Wang, ``Recurrent feature propagation and edge
  skip-connections for automatic abdominal organ segmentation,'' \emph{Expert
  Systems with Applications}, p. 123856, 2024.

\bibitem{lin2021variance}
H.~Lin, Z.~Li, Z.~Yang, and Y.~Wang, ``Variance-aware attention {U}-net for
  multi-organ segmentation,'' \emph{Medical Physics}, vol.~48, no.~12, pp.
  7864--7876, 2021.

\bibitem{huang2023joint}
R.~Huang, Z.~Xu, Y.~Xie, H.~Wu, Z.~Li, Y.~Cui, Y.~Huo, C.~Han, X.~Yang, Z.~Liu,
  and Y.~Wang, ``Joint-phase attention network for breast cancer segmentation
  in {DCE}-{MRI},'' \emph{Expert Systems with Applications}, vol. 224, p.
  119962, 2023.

\bibitem{Hu2018CVPR}
J.~Hu, L.~Shen, and G.~Sun, ``Squeeze-and-excitation networks,'' in
  \emph{Proceedings of the IEEE Conference on Computer Vision and Pattern
  Recognition (CVPR)}, 2018, pp. 7132--7141.

\bibitem{diceloss}
F.~Milletari, N.~Navab, and S.-A. Ahmadi, ``V-net: Fully convolutional neural
  networks for volumetric medical image segmentation,'' in \emph{2016 fourth
  international conference on 3D vision (3DV)}.\hskip 1em plus 0.5em minus
  0.4em\relax Ieee, 2016, pp. 565--571.

\bibitem{wang20203d}
Y.~Wang, C.~Qin, C.~Lin, D.~Lin, M.~Xu, X.~Luo, T.~Wang, A.~Li, and D.~Ni,
  ``3{D} inception {U}-net with asymmetric loss for cancer detection in
  automated breast ultrasound,'' \emph{Medical Physics}, vol.~47, no.~11, pp.
  5582--5591, 2020.

\bibitem{wang2019deeply}
Y.~Wang, N.~Wang, M.~Xu, J.~Yu, C.~Qin, X.~Luo, X.~Yang, T.~Wang, A.~Li, and
  D.~Ni, ``Deeply-supervised networks with threshold loss for cancer detection
  in automated breast ultrasound,'' \emph{IEEE Transactions on Medical
  Imaging}, vol.~39, no.~4, pp. 866--876, 2019.

\bibitem{kervadec2019boundary}
H.~Kervadec, J.~Bouchtiba, C.~Desrosiers, E.~Granger, J.~Dolz, and I.~B. Ayed,
  ``Boundary loss for highly unbalanced segmentation,'' in \emph{International
  Conference on Medical Imaging with Deep Learning}.\hskip 1em plus 0.5em minus
  0.4em\relax PMLR, 2019, pp. 285--296.

\bibitem{unetr}
A.~Hatamizadeh, Y.~Tang, V.~Nath, D.~Yang, A.~Myronenko, B.~Landman, H.~R.
  Roth, and D.~Xu, ``Unetr: Transformers for 3{D} medical image segmentation,''
  in \emph{Proceedings of the IEEE/CVF winter conference on applications of
  computer vision}, 2022, pp. 574--584.

\bibitem{swin}
A.~Hatamizadeh, V.~Nath, Y.~Tang, D.~Yang, H.~R. Roth, and D.~Xu, ``Swin unetr:
  Swin transformers for semantic segmentation of brain tumors in {MRI}
  images,'' in \emph{International MICCAI Brainlesion Workshop}.\hskip 1em plus
  0.5em minus 0.4em\relax Springer, 2021, pp. 272--284.

\bibitem{vit}
A.~Dosovitskiy, L.~Beyer, A.~Kolesnikov, D.~Weissenborn, X.~Zhai,
  T.~Unterthiner, M.~Dehghani, M.~Minderer, G.~Heigold, S.~Gelly, J.~Uszkoreit,
  and N.~Houlsby, ``An image is worth 16x16 words: Transformers for image
  recognition at scale,'' \emph{arXiv preprint arXiv:2010.11929}, 2020.

\end{thebibliography}
\end{document}